\documentclass[manuscript,natbib=false]{acmart}

\AtBeginDocument{%
  }

\setcopyright{acmlicensed}
\copyrightyear{2026}
\acmYear{2026}
\acmDOI{XXXXXXX.XXXXXXX}
\acmConference[HIKM '26]{19th Health Informatics Knowledge Management Conference}{Sept 08--10, 2026}{Australia}
\acmISBN{978-1-4503-XXXX-X/2018/06}

\RequirePackage[
  datamodel=acmdatamodel,
  style=acmnumeric,
  ]{biblatex}

\usepackage[table]{xcolor}
\usepackage{tcolorbox}
\usepackage{caption}

\begin{document}

\title[Evaluating Geographic Sensitivity of Large Language Models in Zero-Shot Chronic Kidney Disease Screening]{Does Country Matter? Evaluating Geographic Sensitivity of Large Language Models in Zero-Shot Chronic Kidney Disease Screening}

\author{Sirajam Munira}
\affiliation{%
  \institution{Rensselaer Polytechnic Institute}
  \city{NY}
  \country{USA}}
\email{munirs@rpi.edu}

\author{Muhammad Ashad Kabir}
\authornote{Corresponding author.}
\affiliation{%
  \institution{Charles Sturt University}
  \city{Bathurst}
  \state{NSW}
  \country{Australia}
}
\email{akabir@csu.edu.au}

\begin{abstract}
  Large language models (LLMs) show strong zero-shot reasoning capabilities, yet their sensitivity to prompt context raises questions about reliability in deployment. This study investigates whether geographic context embedded in prompts influences LLM-based chronic kidney disease (CKD) screening, independent of any change in patient features. Using a community-based CKD dataset from Bangladesh (284 participants with 112 CKD and 172 non-CKD), we systematically varied the country attribute across 25 global contexts while holding all clinical inputs constant, and evaluated two open-weight LLMs (Qwen3 8B and Mistral 7B). Experiments were conducted on a common-correct subset of 156 participants for whom both models produced correct baseline predictions. Results show that non-CKD predictions were highly stable across geographic contexts, with both models maintaining consistent non-CKD predictions in nearly all settings. In contrast, CKD-positive predictions were considerably more sensitive. For Qwen3, CKD prediction dropped from a baseline of 65 to as low as 50 across country contexts, while Mistral showed smaller but non-negligible variation. Country contexts did not group according to known epidemiological or healthcare similarities; instead, groupings appear to reflect how country tokens are internally represented within each model. These findings demonstrate that geographic tokens can systematically shift clinical screening outcomes despite identical patient inputs, with implications for the deployment of LLM-based decision support in diverse healthcare settings.
\end{abstract}





\maketitle

\section{Introduction}
The kidneys play a fundamental role in maintaining physiological homeostasis by filtering metabolic waste, regulating blood pressure, maintaining electrolyte and acid–base balance, and producing hormones essential for red blood cell formation~\cite{lote1994principles}. Chronic kidney disease (CKD) is a progressive condition characterized by sustained decline in kidney function or persistent kidney damage~\cite{kdigo2012}, which can ultimately lead to end-stage renal disease requiring dialysis or kidney transplantation~\cite{kalantar2021chronic}. CKD represents a major global health burden, affecting approximately 850 million people worldwide~\cite{jager2019single}, with a disproportionately high prevalence in low- and middle-income countries where access to early diagnosis and preventive care remains limited~\cite{stanifer2016chronic}. In 2017 alone, CKD accounted for approximately 1.2 million deaths globally and continues to impose substantial clinical and economic burdens on healthcare systems~\cite{bikbov2020global}. 

Early detection of CKD is critical for slowing disease progression and preventing complications such as cardiovascular disease and end-stage kidney failure~\cite{pollock2024framework}. However, early-stage CKD is often asymptomatic, resulting in low patient awareness and delayed diagnosis~\cite{plantinga2008patient}, yet timely detection and intervention can significantly slow disease progression and reduce complications such as cardiovascular morbidity and mortality~\cite{kalantar2021chronic}. Standard diagnostic procedures rely on laboratory biomarkers such as estimated glomerular filtration rate (eGFR) and urine albumin-to-creatinine ratio (ACR)~\cite{levin2024executive}, which require specialized infrastructure and trained personnel, limiting their availability in community and resource-constrained settings~\cite{stanifer2016chronic}. Consequently, there is growing interest in developing computational approaches that can assist in early CKD risk identification using routinely available clinical information.

Over the past decade, machine learning (ML) has been widely explored for CKD prediction and classification. Numerous studies have investigated supervised learning algorithms, including logistic regression, decision trees, random forests, support vector machines, gradient boosting, and neural networks to identify CKD risk from structured clinical variables and electronic health records. Early studies demonstrated the feasibility of using data-driven approaches for kidney disease prediction and patient risk stratification~\cite{kusiak2005predicting}. Subsequent research has reported strong predictive performance across multiple ML models when applied to clinical datasets~\cite{sarker2021community}. Recent systematic reviews further confirm the growing adoption of ML techniques for CKD screening, diagnosis, and risk prediction in both clinical and population-level settings~\cite{kabir2026community,gogoi2025machine, delrue2024application, sanmarchi2023predict}. Collectively, these studies highlight the growing potential of ML-based systems to assist in automated CKD risk assessment and clinical decision support.

Recent advances in large language models (LLMs) have introduced new opportunities for clinical reasoning, decision support, and medical knowledge extraction. Modern LLMs demonstrate strong zero-shot generalization capabilities, enabling them to perform complex reasoning and classification tasks without task-specific training~\cite{kabir2026can,kojima2022large, brown2020language}. Within nephrology, early research on LLM applications has focused primarily on clinical decision support and information extraction tasks, although recent work has begun to explore direct disease screening. Recent reviews highlight the growing role of LLMs in nephrology for tasks such as CKD management guidance, patient education, and knowledge synthesis from clinical literature~\cite{hu2025large, unger2025clinical}. Other studies have explored the use of language models to predict CKD progression from clinical narratives~\cite{awan2025predicting}, assist clinical decision-making in nephrology procedures~\cite{toal2025large}, and extract kidney-related risk factors from patient-reported data~\cite{mao2025identifying}. LLMs have also been used to identify comorbidities and clinical characteristics of chronic diseases through automated analysis of biomedical texts and patient records~\cite{zhang2025extracting, saghy2025novel}. More recently, Kabir and Munira~\cite{kabir2026many} showed that ML-selected features achieved comparable or better zero-shot CKD screening performance than the full feature set. While these studies demonstrate the growing role of LLMs in nephrology and clinical reasoning tasks, their application to structured clinical screening problems is emerging but remains relatively limited. Little is known about how LLMs perform when identifying CKD risk directly from patient attributes in zero-shot settings. Understanding the behavior of LLM-based screening systems under such conditions is therefore an important step toward evaluating their potential for clinical risk assessment.

LLMs are known to be highly sensitive to contextual information in prompts, where even small changes in wording or framing can alter model predictions and probability estimates~\cite{zhao2021calibrate}. This behavior is often linked to broader machine learning phenomena such as dataset shift and contextual priors, where models rely on contextual signals rather than strictly causal relationships between input variables and outcomes~\cite{geirhos2020shortcut, moreno2012unifying}. Modern neural networks are also known to exhibit calibration instability, meaning prediction probabilities can change under contextual or distributional shifts~\cite{guo2017calibration}. In clinical settings, such sensitivity raises important concerns, as LLM-based diagnostic reasoning can vary depending on how clinical scenarios are framed in prompts~\cite{singhal2023large}. Prior research further shows that demographic or geographic attributes embedded in prompts can influence model outputs even when the underlying information remains unchanged~\cite{deshpande2023toxicity, naous2025origin, santurkar2023whose}.

Despite growing evidence of contextual sensitivity in language models, the influence of geographic context on clinical inference remains largely unexplored. In healthcare, disease prevalence, diagnostic practices, and healthcare infrastructure vary substantially across regions, and such contextual information may implicitly affect model predictions when included in prompts. Consequently, a country attribute embedded within a prompt may alter the model’s decision boundary even when the patient's clinical characteristics remain unchanged. This issue is particularly relevant for screening tasks such as CKD detection, where consistent and reliable predictions are essential for population-level risk assessment. However, there has been limited empirical investigation into whether geographic tokens influence LLM-based predictions in clinical screening scenarios.

To address this gap, we investigate whether modifying only the country attribute in a clinical prompt can influence LLM-based CKD screening outcomes while all patient features remain unchanged. Using a controlled prompt intervention framework, we evaluate prediction stability across multiple geographic contexts and quantify the resulting changes in model behavior. Beyond CKD screening, this study contributes to the broader goal of trustworthy healthcare AI by examining whether non-clinical contextual attributes can influence clinical inference despite identical clinical evidence. Understanding such effects is essential for the reliable deployment of LLM-based decision-support systems across diverse populations and settings.

\section{Methodology}
\begin{figure*}[!ht]
    \centering
    \includegraphics[width=1\linewidth]{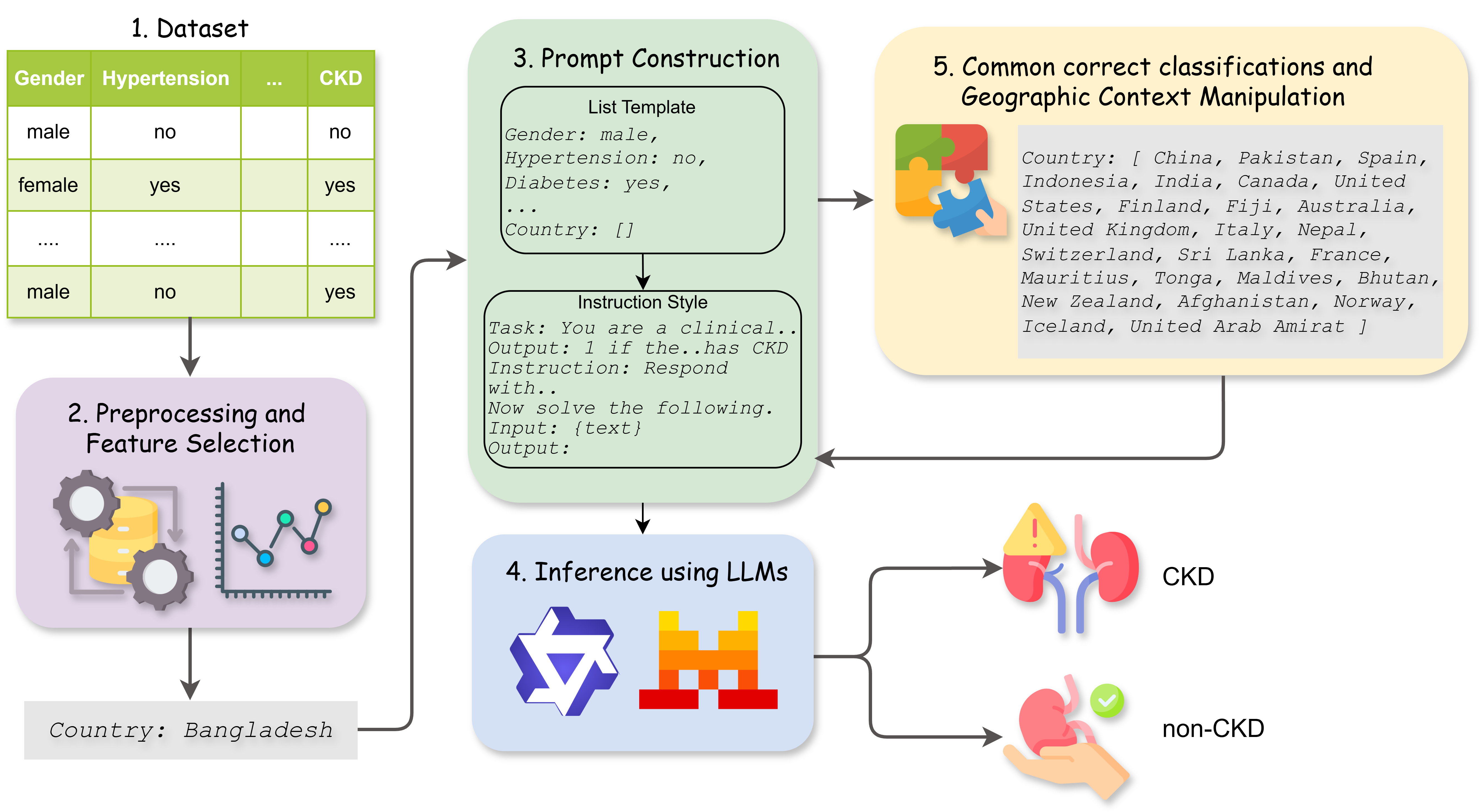}
    \caption{Schematic overview of the zero-shot LLM pipeline for evaluating geographic sensitivity in CKD screening.}
    \label{fig:methodology}
\end{figure*}

Figure~\ref{fig:methodology} provides an overview of the proposed zero-shot LLM pipeline for evaluating geographic sensitivity in CKD screening. The workflow consists of four main steps: (i) data preprocessing and feature selection, (ii) prompt construction, (iii) inference using large language models, and (iv) geographic context manipulation and evaluation.

\subsection{Dataset}
We evaluate the proposed framework using a community-based CKD dataset~\cite{sarker2021community} to determine the prevalence, health, and nutritional status of patients with chronic kidney disease in Bangladesh. The dataset was collected from the Mirzapur sub-district of Tangail, Bangladesh, and represents one of the earliest community cohorts designed for early-stage CKD screening in a South Asian population. Participants were recruited through age-stratified random sampling of adults aged 18 years and above within the Mirzapur demographic surveillance system. Data collection involved trained community health workers who conducted structured household interviews, physical examinations, and laboratory assessments. Blood and urine samples were obtained to measure key clinical biomarkers, including serum creatinine, fasting glucose, lipid profile indicators, hemoglobin levels, and urine albumin-to-creatinine ratio (ACR). CKD diagnosis was established based on sustained reductions in estimated glomerular filtration rate (eGFR) and/or elevated ACR following established clinical diagnostic guidelines.

After preprocessing to remove incomplete records, the final analytic dataset consisted of 284 participants, including 112 early-stage CKD cases (stages 1–3) and 172 non-CKD controls. The dataset contains 24 categorical variables spanning multiple domains relevant to CKD risk assessment. These include socio-demographic characteristics (e.g., age group, gender, literacy status, occupation, marital status), lifestyle factors (e.g., sleep duration, smoking, smokeless tobacco use), medical history indicators (e.g., hypertension, diabetes, heart disease, stroke, and family history of hypertension), clinical examination variables (e.g., body mass index category, abdominal obesity, undernutrition, anemia), and pathological findings (e.g., presence of red blood cells in urine, low serum albumin, and dyslipidemia markers). Together, these variables capture a multidimensional risk profile reflecting demographic, behavioral, and clinical characteristics associated with early CKD detection in community-based screening settings.

\subsection{Data Preprocessing and Feature Selection}
Data preprocessing and feature selection for this study follow the methodology established in the prior study~\cite{kabir2026community} on machine learning analysis of the same CKD dataset. In that work, standard preprocessing steps were applied to ensure data consistency and analytical readiness. Specifically, categorical variables were harmonized and represented in a structured format suitable for computational modeling, while continuous variables were discretized into categorical groups where appropriate to align with the dataset’s design. These steps ensured that the features remained interpretable and consistent with the community-based screening context. 

Feature selection was conducted in the previous study~\cite{kabir2026community} using a comprehensive framework that combined multiple statistical, regularization-based, and wrapper-based techniques to identify the most informative predictors for early CKD detection. Based on the outcomes of that analysis, a consolidated subset of clinically relevant variables was identified. In this study, we directly adopt this previously validated feature subset as the input representation for the large language models in the zero-shot screening framework. The selected predictors include hypertension history, age group (60+ years), diabetes history, anemia status, obesity (BMI category), presence of red blood cells in urine (RBC), daily sleep duration less than seven hours, gender, and family history of hypertension. These features capture a combination of demographic characteristics, lifestyle indicators, medical history, and clinical findings relevant to early CKD risk assessment. Using this compact and clinically meaningful subset allows the LLM-based screening framework to focus on the most informative predictors identified through prior machine learning experiments. Table~\ref{tab:datasetFeatures} summarizes the selected feature subset used in this study.

\begin{table}[!t]
    \centering
        \caption{All features and selected features (underlined) of the dataset.}
    \label{tab:datasetFeatures}
    \begin{tabular}{p{\linewidth}}
    \hline
       All features\\
       \hline\hline
        \underline{Age}, \underline{Gender}, Illiterate, Occupation, Marital status, \underline{Sleeping duration}, Tobacco smoker, Smokeless tobacco, \underline{History of hypertension}, \underline{History of diabetes}, Heart disease, Stroke, Family history of diabetes, \underline{Family history of hypertension}, Family history of CKD, \underline{Body mass index}, Abdominal obesity, Undernutrition, \underline{Anemia}, \underline{Presence of red blood cells in urine}, Serum albumin, Hyper-cholesterolemia, HDL cholesterol, Hypertriglyceridemia\\
        \hline
        
        \hline
    \end{tabular}
\end{table}

\subsection{Prompt Construction}
To enable zero-shot inference with large language models, the selected patient features were incorporated into the prompt as feature-value pairs, where each attribute and its corresponding value were explicitly listed in a structured format. This representation allows the language models to interpret individual clinical attributes while preserving the semantics of the original tabular data~\cite{hegselmann2023tabllm}.

We adopt an instruction-style prompting strategy in which the patient feature list is embedded within a task-oriented instruction: input-output format~\cite{wang2023self}. The prompt explicitly defines the clinical classification objective, instructing the model to determine whether a patient has chronic kidney disease (CKD) based on the provided features. The model is assigned to the role of a clinical classification system and is constrained to produce a binary output, where 1 indicates CKD and 0 indicates non-CKD. To reduce response variability and ensure consistent outputs, the prompt specifies that the model must return exactly one token corresponding to the predicted class label. This instruction-based prompt structure was chosen because it aligns with instruction-tuning paradigms commonly used during LLM training, enabling effective zero-shot generalization across different models~\cite{wei2021finetuned}. In addition, the instruction-style format avoids reliance on model-specific chat tokens, improving prompt compatibility and consistency when evaluating multiple LLM architectures. An example of the prompt template used in this study is presented in Table~\ref{tab:prompt_templates}.

\begin{table}[ht]
\centering
\caption{Example of the instruction-style prompt used for zero-shot inference.}
\label{tab:prompt_templates}

\begin{tcolorbox}
{\scriptsize
Task: You are a clinical classification model. Classify whether a patient has chronic kidney disease (CKD) based on the patient's input features. 1 indicates CKD, and 0 indicates non-CKD.

Output 1 if the patient has CKD.

Output 0 if the patient does not have CKD.

Instruction: Respond with exactly one token: 0 or 1.

Now solve the following.

Input:

Patient features:

History of diabetes: \textit{Yes},

History of hypertension: \textit{No},

...

Output:
}
\end{tcolorbox}

\end{table}

\subsection{Large Language Models Used for Inference}
In this study, we evaluated two large language models, Qwen3-8B~\cite{qwen3technicalreport} and Mistral 7B~\cite{jiang2023mistral7b}, to investigate the sensitivity of LLM-based clinical screening to geographic context. These models were selected due to their strong general-purpose reasoning capabilities and widespread use in recent LLM evaluation studies. In addition, they represent two widely used open-weight LLM families with different architectures and training pipelines, allowing us to examine whether geographic-context sensitivity can be observed across more than one model rather than within a single system. Both models are pretrained on large-scale heterogeneous corpora that include medical and scientific text, enabling them to perform zero-shot reasoning on healthcare-related tasks without task-specific training.

Inference was conducted in a zero-shot setting, where the models were provided with the constructed instruction-style prompt containing the patient feature list and asked to predict whether the patient has chronic kidney disease (CKD). The output space was explicitly constrained to a binary decision, where 1 indicates CKD and 0 indicates non-CKD, ensuring consistent and comparable outputs across models. Each patient record was evaluated independently, and no in-context examples or parameter updates were used. This setup allows us to directly examine how the models perform clinical screening based solely on the provided patient features and contextual information, including the geographic attribute introduced in the prompt.

\subsection{Geographic Context Manipulation and Evaluation}
To examine whether large language models adjust their screening behavior based on geographic context, we conducted a controlled prompt intervention experiment. First, zero-shot inference was performed on the full dataset using both Qwen3 and Mistral. From these predictions, we identified the subset of participants that were correctly classified by both models, forming a common-correct subset. Restricting the analysis to this subset ensured that all evaluated cases had correct baseline predictions under the original Bangladesh context. This design isolates geographic sensitivity from baseline model error, allowing observed prediction changes to be attributed more directly to contextual variation rather than pre-existing misclassification noise.

For each participant in this subset, the exact same prompt structure and patient feature values were retained, and only the country attribute describing the patient’s geographic context was modified as a controlled contextual variable. No other elements of the prompt were changed. The modified prompts were then re-evaluated using the same inference setup for both models. 
A total of 25 alternative country contexts were evaluated in addition to the original Bangladesh context, resulting in 26 geographic settings overall. The evaluated geographic contexts included Bangladesh, China, Pakistan, Spain, Indonesia, India, Canada, the United States, Finland, Fiji, Australia, the United Kingdom, Italy, Nepal, Switzerland, Sri Lanka, France, Mauritius, Tonga, Maldives, Bhutan, New Zealand, Afghanistan, Norway, Iceland, and the United Arab Emirates. The selected countries were chosen to represent diverse geographic regions, income levels, healthcare systems, population sizes, and CKD prevalence profiles, including countries from North America, Europe, Asia, Oceania, and island nations. The objective was not to model country-specific CKD epidemiology but rather to assess whether geographic tokens themselves influence LLM predictions across varied contextual settings.

To assess whether changes in prediction outcomes were statistically significant relative to the Bangladesh baseline, we performed McNemar's test for each country-specific context. The test compared the number of CKD cases that changed classification after modifying the country attribute while keeping all patient features unchanged. Statistical significance was evaluated at the 0.05 level.

By comparing the predictions obtained before and after changing the country attribute, we assessed whether the models’ classification outcomes remained stable or shifted when only the geographic context was altered. This controlled design isolates the effect of geographic context while holding all clinical variables constant and directly evaluates whether contextual attributes can affect model behavior despite identical clinical inputs.

\section{Results}

\subsection{Baseline performance and common-correct subset}
We first examined model predictions on the original dataset to establish a stable reference set for the geographic-context analysis. From the initial zero-shot inference results, we identified the subset of participants for whom both models produced correct predictions for the actual geographical context, i.e., when country = Bangladesh. The common-correct subset consisted of 156 participants (65 CKD, 91 non-CKD), which served as the reference population for all subsequent experiments. Because these participants were consistently classified correctly by both models under the original context, they provide a controlled set of instances where baseline model agreement is high. 

This subset serves as the baseline population for evaluating whether changes in geographic context can influence model behavior when the underlying clinical information remains unchanged. Subsequent experiments, therefore, assess whether modifying only the country attribute in the prompt leads to changes in predictions within this subset.

\begin{figure*}[!ht]
    \centering
    \includegraphics[width=1\linewidth]{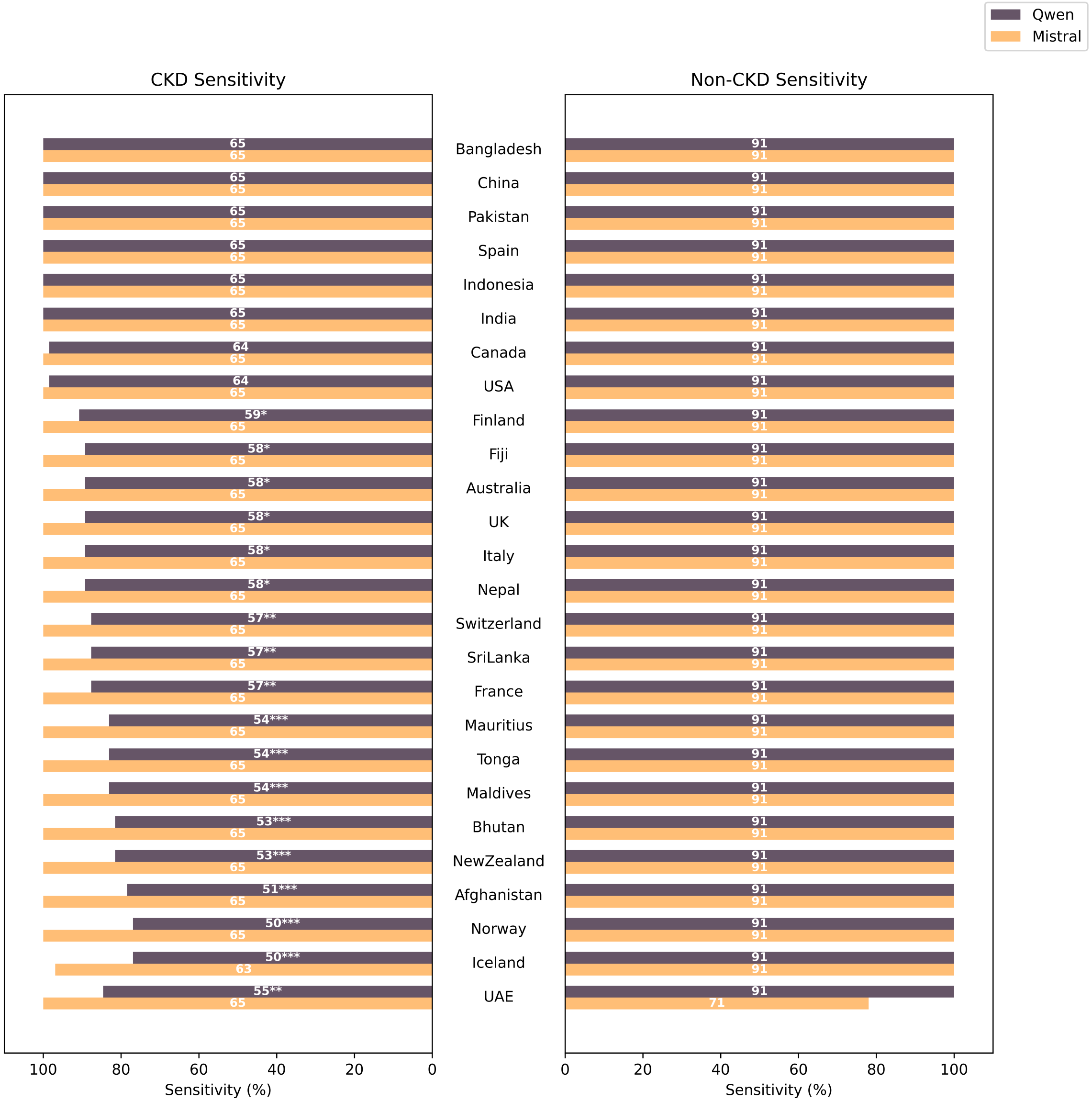}
    \caption{CKD sensitivity and non-CKD sensitivity across geographic contexts for Qwen and Mistral. Numbers inside bars indicate correctly identified cases. Statistical significance relative to the Bangladesh baseline is denoted by McNemar test * ($p<0.05$), ** ($p<0.01$), and *** ($p<0.001$).}
    \label{fig:result1}
\end{figure*}

\subsection{Prediction stability across geographic contexts}
Figure~\ref{fig:result1} presents CKD sensitivity and non-CKD sensitivity across 26 geographic contexts, including the original Bangladesh context and 25 additional countries spanning North America (Canada, USA), Europe (Spain, Finland, United Kingdom, Italy, Switzerland, France, Norway, Iceland), Asia (China, Pakistan, Indonesia, India, Nepal, Sri Lanka, Bhutan, Afghanistan, United Arab Emirates), Oceania (Australia and New Zealand), and island nations (Fiji, Mauritius, Tonga, and Maldives). These countries were selected to represent diverse geographic regions, income levels, healthcare systems, population sizes, and CKD prevalence profiles.

The evaluation was conducted on the common-correct subset consisting of 65 CKD cases and 91 non-CKD cases. For the non-CKD cases, Qwen3 maintained a non-CKD sensitivity of 100\%, consistently identifying all 91 cases correctly across all 25 geographic contexts. Mistral also maintained 100\% non-CKD sensitivity for nearly all countries. However, when the country attribute was changed to the United Arab Emirates (UAE), the number of correctly identified non-CKD cases decreased from 91 to 71, representing a non-CKD sensitivity decrease to 78\%.

For CKD cases, the effects of geographic context were more pronounced. Under the Bangladesh baseline, both models achieved a CKD sensitivity of 100\%. Qwen maintained identical performance for China, Pakistan, Spain, Indonesia, and India. However, CKD sensitivity declined across several other contexts, reaching 90.8\% for Finland, 87.7\% for Switzerland, Sri Lanka, and France, and 76.9\% for Norway and Iceland. 
Many of these reductions were statistically significant relative to the Bangladesh baseline (e.g., Finland, Fiji, Australia, the United Kingdom, Italy, Nepal, Switzerland, Sri Lanka, France, the United Arab Emirates, Mauritius, Tonga, Maldives, Bhutan, New Zealand, Afghanistan, Norway, and Iceland), with significance increasing as the sensitivity decrease became larger. In contrast, Mistral demonstrated substantially greater stability, maintaining 100\% CKD sensitivity across all contexts except Iceland, where sensitivity decreased slightly to 96.9\%, a change that was not statistically significant.

Overall, geographic context had little influence on most non-CKD predictions but produced measurable changes in CKD detection performance. The effects were not uniform across countries or models. While some countries, including China, Pakistan, Spain, Indonesia, and India, produced results identical to the Bangladesh baseline, others, including Norway, Iceland, and the UAE, resulted in substantially different prediction outcomes despite identical patient characteristics. These findings suggest that country attributes can act as contextual signals that influence LLM-based clinical inference, although the magnitude and direction of the effect vary across models and geographic settings.

\subsection{Model-specific sensitivity}
Although both LLMs exhibited sensitivity to geographic context, the magnitude of this effect differed substantially between models. Qwen showed a broad range of CKD sensitivities across countries, varying from 100\% under several contexts to 76.9\% for Norway and Iceland. Statistically significant reductions relative to the Bangladesh baseline were observed across multiple countries, including Finland ($p<0.05$), Switzerland, Sri Lanka, France, and the UAE ($p<0.01$), as well as Mauritius, Tonga, Maldives, Bhutan, New Zealand, Afghanistan, Norway, and Iceland ($p<0.001$).

In contrast, Mistral demonstrated markedly greater stability in CKD detection. CKD sensitivity remained at 100\% for all evaluated countries except Iceland, where sensitivity decreased slightly to 96.9\% (63/65). This reduction was not statistically significant. For non-CKD detection, both models maintained perfect sensitivity across nearly all contexts, although Mistral exhibited a notable exception for the UAE, where non-CKD sensitivity decreased to 78.0\%.

Overall, these findings indicate that geographic context affects the two models differently. Qwen exhibits broader and statistically significant variation in CKD sensitivity across countries, whereas Mistral remains largely stable with only isolated deviations. This suggests that contextual sensitivity may depend on model architecture, pretraining data, or internal representations of geographic information.

\subsection{Geographic effects}
Beyond overall prediction stability, the country-level analysis reveals several systematic patterns in how geographic context influences model behavior. Here, the baseline refers to the original Bangladesh context used in the source dataset. When the same patient features are evaluated under different country attributes, the resulting predictions do not vary randomly; instead, multiple countries produce similar prediction outcomes. The country-level observations are presented descriptively to highlight similarities in prediction patterns across contexts; no formal clustering algorithm was applied.

The most stable results were observed for Canada and the United States. Under these contexts, both models produced prediction outcomes that remained nearly identical to the Bangladesh baseline. For Qwen3, all 65 CKD cases and all 91 non-CKD cases were correctly identified, while Mistral maintained 65 correctly identified CKD cases and 91 correctly identified non-CKD cases. These results indicate minimal sensitivity to geographic context for these two countries.

A second pattern was observed for Finland, Fiji, Australia, the United Kingdom, Italy, Nepal, Switzerland, Sri Lanka, and France. Across these contexts, Qwen consistently exhibited moderate reductions in CKD sensitivity ranging from 87.7\% to 90.8\%. Several of these reductions were statistically significant relative to the Bangladesh baseline. Mistral remained largely unchanged across the same countries, maintaining a CKD sensitivity of 100\%.

A third pattern emerged for Mauritius, Tonga, Maldives, Bhutan, New Zealand, and the United Arab Emirates. Across these contexts, Qwen exhibited larger reductions in CKD sensitivity, ranging from 81.5\% to 84.6\%, with all reductions reaching statistical significance. Within this group, prediction outcomes remained highly similar across countries, indicating stronger contextual effects associated with these geographic attributes.

The largest deviations from the Bangladesh baseline were observed for Afghanistan, Norway, and Iceland. For Qwen, CKD sensitivity decreased to 78.5\% for Afghanistan and 76.9\% for Norway and Iceland. All three reductions were statistically significant ($p<0.001$). These countries represent substantially different geographic, economic, and healthcare contexts, suggesting that the observed prediction shifts are unlikely to be driven by real-world similarities between countries.

When these patterns are compared with global epidemiological statistics on CKD prevalence~\cite{bikbov2020global}, no consistent alignment emerges. For example, countries such as Tonga, Mauritius, and Maldives are associated with relatively high CKD prevalence, whereas Finland, Italy, France, Norway, and Iceland are among countries with substantially lower prevalence estimates~\cite{bikbov2020global}. However, countries with similar prevalence profiles did not consistently produce similar prediction outcomes, and countries with very different prevalence profiles often resulted in nearly identical model behavior. These findings suggest that the observed prediction shifts are unlikely to reflect real-world CKD epidemiology.

Another notable observation is that prediction changes were concentrated among a relatively small subset of CKD-positive participants. Several participants who were correctly classified under the Bangladesh baseline became misclassified across multiple country contexts, while the majority of non-CKD participants remained correctly identified across nearly all evaluated countries. This indicates that geographic context primarily influences a subset of borderline cases rather than uniformly affecting all predictions.

Overall, the results demonstrate that changing only the country attribute can systematically alter zero-shot CKD screening outcomes. Similar prediction patterns were observed for groups of countries including Canada and the United States; Finland, Fiji, Australia, the United Kingdom, Italy, Nepal, Switzerland, Sri Lanka, and France; Mauritius, Tonga, Maldives, Bhutan, New Zealand, and the United Arab Emirates; and Afghanistan, Norway, and Iceland. Because these groups do not correspond closely to geographic proximity, healthcare system similarity, or CKD prevalence patterns, the findings suggest that country tokens function as contextual signals within the models, influencing clinical inference despite identical patient characteristics.

\section{Discussion}
This study examined whether geographic context embedded in prompts influences large language model (LLM) predictions for chronic kidney disease (CKD) screening. The results show that modifying only the country attribute, while keeping all patient features unchanged, can lead to observable shifts in prediction outcomes. These effects were most visible in the CKD-positive cases, where the number of correctly identified cases varied across geographic contexts. In contrast, predictions for non-CKD cases remained largely stable across most countries. 

The analysis further revealed that countries tend to produce similar groups of prediction outcomes. Certain contexts, such as the United States and Canada, generated results that remained very close to the Bangladesh baseline, whereas countries including Mauritius, Tonga, Maldives, Bhutan, New Zealand, Afghanistan, Norway, and Iceland produced larger deviations. Importantly, these groupings did not align consistently with known epidemiological or healthcare similarities between countries. This suggests that geographic tokens may function as contextual signals within the models rather than reflecting real-world medical patterns. Differences were also observed between the evaluated models. Qwen3 exhibited greater variation in CKD-positive predictions across geographic contexts, whereas Mistral showed generally smaller fluctuations with a few isolated deviations.

From a trustworthy AI perspective, these findings highlight a previously underexplored source of instability in LLM-based clinical screening systems. The observed prediction shifts occurred despite identical patient characteristics, suggesting that non-clinical contextual attributes may influence model behavior during inference. As LLMs continue to be explored for healthcare applications, systematic evaluation of contextual sensitivity will be important for ensuring reliable and consistent clinical decision support across diverse deployment settings.

While the present experiments cannot identify the underlying mechanisms, several plausible explanations may account for the observed effects. Large language models are trained on large-scale internet and text corpora that contain information associated with countries, populations, healthcare systems, and disease-related narratives. Consequently, country tokens may activate latent contextual representations that influence the inference process. Geographic attributes may also function as contextual priors, causing models to associate certain countries with different disease risks or healthcare-related patterns. More broadly, the observed behavior suggests that contextual cues embedded in prompts can influence clinical inference even when the underlying clinical information remains unchanged.

\section{Limitations and Future Work}
This study has several limitations. First, the experiments were conducted using a single community-based CKD dataset from Bangladesh, which may limit the generalizability of the findings to other populations and diseases. Second, only two open-weight LLMs were evaluated; therefore, the observed effects should not be assumed to represent the behavior of all model families. Third, geographic context was represented solely through country attributes, whereas real-world clinical contexts may involve additional demographic, cultural, and healthcare-related factors.

Future work should evaluate larger and more diverse clinical datasets, examine additional open-weight and proprietary LLMs, and investigate other contextual attributes that may influence model predictions. Further research is also needed to better understand how contextual information is encoded within LLMs and how such effects can be mitigated to improve the reliability and robustness of LLM-assisted clinical decision support systems.

\section{Conclusion}
This study investigated whether geographic context influences large language model predictions in zero-shot chronic kidney disease (CKD) screening. Using a controlled prompt intervention framework, we showed that modifying only the country attribute can alter screening outcomes despite identical patient characteristics. The observed effects varied across countries and models, indicating that geographic information can function as a contextual signal during clinical inference.
These findings highlight an important consideration for the use of LLMs in healthcare: non-clinical prompt attributes may influence model predictions even when the underlying clinical evidence remains unchanged. As LLM-based decision-support systems continue to be explored for medical applications, understanding and evaluating contextual sensitivity will be essential for ensuring reliable, robust, and trustworthy clinical AI systems across diverse deployment settings.

\printbibliography


\end{document}